\pdfoutput=1

\documentclass[11pt]{article}

\usepackage[final]{coling}

\usepackage{times}
\usepackage{latexsym}

\usepackage[T1]{fontenc}

\usepackage[utf8]{inputenc}

\usepackage{microtype}

\usepackage{inconsolata}

\usepackage{graphicx}
\usepackage{tcolorbox}
\usepackage{booktabs}
\usepackage{multirow}
\usepackage{arydshln}

\title{Leveraging Open-Source Large Language Models for Native Language Identification}

\author{Yee Man Ng \\
  CLTL, Vrije Universiteit Amsterdam \\
  Amsterdam, The Netherlands \\
  \texttt{y.m.ng@student.vu.nl}\\\And
  Ilia Markov \\
  CLTL, Vrije Universiteit Amsterdam \\
  Amsterdam, The Netherlands \\
  \texttt{i.markov@vu.nl} \\}

\begin{document}
\maketitle
\begin{abstract}
Native Language Identification (NLI) -- the task of identifying the native language (L1) of a person based on their writing in the second language (L2) -- has applications in forensics, marketing, and second language acquisition. 
Historically, conventional machine learning approaches that heavily rely on extensive feature engineering have outperformed transformer-based language models on this task.
Recently, closed-source generative large language models (LLMs), e.g., GPT-4, have demonstrated remarkable performance on NLI in a zero-shot setting, including promising results in open-set classification. 
However, closed-source LLMs 
have many disadvantages, such as high costs and undisclosed nature of training data. 
This study explores the potential of using open-source LLMs for NLI. 
Our results indicate that open-source LLMs do not reach the accuracy levels of closed-source LLMs when used out-of-the-box. 
However, when fine-tuned on labeled training data, open-source LLMs can achieve performance comparable to that of commercial LLMs. 
\end{abstract}

\section{Introduction} \label{intro}
Native Language Identification (NLI) is the task of automatically identifying an author's native language (L1) based on texts written in their second language (L2). The task is based on the language transfer hypothesis, the phenomenon in which 
characteristics of L1 influence the  production of texts in L2 to the degree that L1 is identifiable~\cite{Odlin:89}. 
NLI is useful for educational purposes, forensic applications in the context of author profiling, and to inform second language acquisition research~\citep{goswami-etal-2024-native}. 

From a machine learning (ML) perspective, NLI is commonly framed as a supervised multi-class classification task, where NLI systems are trained to assign an author's L1. 
While the task has been proven difficult to perform by humans \citep{malmasi-etal-2015-oracle}, automated methods have shown remarkable results  
using conventional ML approaches based on extensive feature engineering, e.g.,~\cite{cimino-dellorletta-2017-stacked, Markov:2018}. Such methods rely on features that capture L1-indicative linguistic patterns in L2 writing, 
e.g., spelling errors \citep{koppel_2005, chen-etal-2017-improving, markov-etal-2019-anglicized}, word choice \citep{brooke-hirst-2012-robust}, and syntactic patterns \citep{wong-dras-2011-exploiting}. 

Transformer-based encoder models, like BERT~\cite{devlin:2019-bert}, on the other hand, 
have yielded poorer performance than conventional ML approaches for the NLI task \citep{markov-exploiting2022, steinbakken-gamback-2020-native, goswami-etal-2024-native}. 
Previous research suggests that this is likely because NLI concerns very specific linguistic features that models trained on general corpora cannot capture~\citep{ markov-exploiting2022}.
Recent research has shown that generative large language models (LLMs) demonstrate promising results for NLI. \citet{lotfi-etal-2020-deep} presented the first study addressing NLI using fine-tuned GPT-2 models, which outperformed previous traditional ML approaches and achieved state-of-the-art results on the NLI benchmark TOEFL11 and ICLE datasets. 
\citet{zhang-2023} explored the ability of GPT-3.5 \citep{gpt3-brown:20} and GPT-4 \citep{openai-gpt4:23} to perform NLI. Their results indicate that out-of-the-box GPT models demonstrate outstanding performance, with GPT-4 setting a new performance record of 91.7\% accuracy on the TOEFL11 benchmark dataset, and achieve promising results for open-set classification (without a predefined set of L1s), a useful setting for real-world NLI applications. 

While \citeauthor{zhang-2023}'s results indicate that LLMs achieve state-of-the-art performance on NLI, they only evaluate the performance of GPT-3.5 and GPT-4. The closed-source nature of these models presents a multitude of limitations to research. Providers of closed-source models often disclose  minimal information regarding the training data or procedure, hindering the evaluation of results achieved with these models and obscuring biases in training data and models \citep{balloccu-etal-2024-leak}. 
The undisclosed nature of the training data has also raised concerns among researchers about data contamination risks, as it is challenging to determine whether a model's high performance on a task can be attributed to the model's effective generalization or potential data leakage \citep{yu-open-2023}. 
In addition, closed-source models are typically only accessible via an API, causing lack of control over model updates, 
which are often communicated poorly to users \citep{yu-open-2023, pozzobon-etal-2023-challenges}. In turn, the reproducibility of experiments  
cannot be guaranteed. The usage of closed-source LLMs is also highly costly, which negatively impacts the accessibility of LLMs~\citep{bender2021}. 

Providers of open-source LLMs, on the other hand, often release more information regarding training data and procedures. 
As model weights are released openly, open-source LLMs can be fine-tuned for a down-stream task, which is often highly costly or not supported for closed-source models. 
Despite these advantages, employing open-source LLMs for NLI remains unexplored, and it is therefore important to investigate the difference in performance between open-source and proprietary LLMs on this task. 
Hence, the research question addressed in this study is: \textit{Can open-source LLMs be used for effective Native Language Identification?} 

The contributions of this work are the following: (i) we are the first to explore the performance of open-source LLMs on NLI and quantify the difference in performance with closed-source models, and (ii) 
we investigate the impact of fine-tuning open-source LLMs on NLI performance.

\section{Data and Models} \label{method-data} 

To comprehensively evaluate the ability of current LLMs to perform NLI, we compare the performance of two closed-source commercial LLMs (i.e., GPT-3.5 and GPT-4) with five open-source LLMs (\S{\ref{sec:models}), used out-of-the-box and after fine-tuning, on two NLI benchmark datasets. 

\subsection{Data} \label{data}

\paragraph{TOEFL11} ~\cite{Blanchard:13}:
the ETS Corpus of Non-Native Written English (TOEFL11) consists of 12,100 essays, with 1,100 essays per L1, written by English learners with low, medium, or high proficiency levels. The 11 L1s covered in the data are Arabic (ARA), Chinese (CHI), French (FRE), German (GER), Hindi (HIN), Italian (ITA), Japanese (JPN), Korean (KOR), Spanish (SPA), Telugu (TEL), and Turkish (TUR).
We use the TOEFL11 test set for evaluation, which contains 100 essays per L1. 
The average length of essays in TOEFL11 is 348 words. 

\paragraph{ICLE-NLI}~\cite{Granger:09}:
a 7-language subset of the ICLEv2 dataset commonly used for NLI~\cite{Tetreault:12}. The data contains 770 essays, with 110 essays per L1, written by highly-proficient English learners. The L1s represented in the dataset are Bulgarian (BUL), Chinese (CHI), Czech (CZE), French (FRE), Japanese (JPN), Russian (RUS), and Spanish (SPA). We evaluate the models on the complete ICLE-NLI dataset. The average length of essays in this corpus is 747 words.  

\begin{table*}[htb!]
\centering
\begin{tabular}{lcccc}
\toprule
\multirow{2}{*}{\textbf{Model}} & \multicolumn{2}{c}{\begin{tabular}[c]{@{}c@{}}\textbf{TOEFL11} \\ (11 L1s, test set)\end{tabular}} & \multicolumn{2}{c}{\begin{tabular}[c]{@{}c@{}}\textbf{ICLE-NLI}\\ (7 L1s, 5FCV/entire)\end{tabular}} \\
                                & \multicolumn{1}{l}{Closed-set}                & \multicolumn{1}{l}{Open-set}               & \multicolumn{1}{l}{Closed-set}                 & \multicolumn{1}{l}{Open-set}                 \\ \midrule
\textit{Baselines}              & \multicolumn{1}{l}{}                          & \multicolumn{1}{l}{}                       & \multicolumn{1}{l}{}                           & \multicolumn{1}{l}{}                         \\
BoW SVM~\citep{lotfi-etal-2020-deep} & 71.1 & -- & 80.6 & -- \\
Feature-engineered  SVM~\citep{Markov:2018}                      & 88.6                                          & --                                         & 93.4                                           & --                                           \\
BERT~\citep{lotfi-etal-2020-deep}                             & 80.8                                          & --                                         & 76.8                                           & --                                           \\
GPT-2 (fine-tuned)~\citep{lotfi-etal-2020-deep}                     & 89.0                                          & --                                         & 94.2                                           & --                                           \\
GPT-3.5~\citep{zhang-2023}                   & 74.0                                          & 73.4                                       & 81.2                                          & 84.2                                        \\
GPT-4~\citep{zhang-2023}                     & \textbf{91.7}                                          & \textbf{86.7}                                       & 95.5                                          & \textbf{89.1}                                        \\ \midrule
\textit{Open-source LLMs}       & \multicolumn{1}{l}{}                          & \multicolumn{1}{l}{}                       & \multicolumn{1}{l}{}                           & \multicolumn{1}{l}{}                         \\
LLaMA-2 (7B) (zero-shot)                        & 29.2 \textcolor{gray}{\footnotesize{±0.9}}                                         & 22.1 \textcolor{gray}{\footnotesize{±0.7}}                                     & 29.2 \textcolor{gray}{\footnotesize{±1.0}}                                       & 15.5 \textcolor{gray}{\footnotesize{±0.3}}                                      \\
LLaMA-2 (7B) (fine-tuned)                    & 78.7 \textcolor{gray}{\footnotesize{±1.0}}                                        & --                                         & 42.9 \textcolor{gray}{\footnotesize{±2.0}}                                        & --                                     \\ \hdashline
LLaMA-3 (8B) (zero-shot)                        & 56.8 \textcolor{gray}{\footnotesize{±1.1}}                                       & 56.4 \textcolor{gray}{\footnotesize{±0.7}}                              & 75.8 \textcolor{gray}{\footnotesize{±0.4}}                                      & 71.0 \textcolor{gray}{\footnotesize{±0.9}}                                      \\
LLaMA-3 (8B) (fine-tuned)                    & 85.3 \textcolor{gray}{\footnotesize{±0.1}}                                    & --                                         & 78.5 \textcolor{gray}{\footnotesize{±2.5}}                                       & --                                           \\
\hdashline
Gemma (7B) (zero-shot)                          & 13.6 \textcolor{gray}{\footnotesize{±0.0}}                                       & 7.0 \textcolor{gray}{\footnotesize{±0.0}}                              & 28.2 \textcolor{gray}{\footnotesize{±0.1}}                                   & 13.1 \textcolor{gray}{\footnotesize{±0.0}}                                       \\
Gemma (7B) (fine-tuned)                      & 90.3 \textcolor{gray}{\footnotesize{±1.2}}                              & --                                         & \textbf{96.6} \textcolor{gray}{\footnotesize{±0.2}}                                & --                                           \\
\hdashline
Mistral (7B) (zero-shot)                        & 35.6 \textcolor{gray}{\footnotesize{±1.6}}                                   & 24.2 \textcolor{gray}{\footnotesize{±0.1}}                               & 53.1 \textcolor{gray}{\footnotesize{±1.1}}                                       & 41.5 \textcolor{gray}{\footnotesize{±0.1}}                                      \\
Mistral (7B) (fine-tuned)                    & 89.8 \textcolor{gray}{\footnotesize{±0.8}}                                    & --                                         & 83.2 \textcolor{gray}{\footnotesize{±9.4}}                                     & --                                           \\
\hdashline
Phi-3 (3.8B) (zero-shot)                          & 18.2 \textcolor{gray}{\footnotesize{±0.3}}                                     & 21.6 \textcolor{gray}{\footnotesize{±1.6}}                                & 33.6 \textcolor{gray}{\footnotesize{±0.4}}                                     & 40.9 \textcolor{gray}{\footnotesize{±2.1}}                                     \\
Phi-3 (3.8B) (fine-tuned)                      & 65.6 \textcolor{gray}{\footnotesize{±0.4}}                                   & --                                         & 51.4 \textcolor{gray}{\footnotesize{±1.7}}                                  & --                                           \\ \bottomrule
\end{tabular}
\caption{Comparative analysis of the performance of the baseline methods and closed- and open-source LLMs on the TOEFL11 and ICLE-NLI datasets in terms of classification accuracy (\%).}
\label{tab:results}
\end{table*}

\subsection{Models}
\label{sec:models}
\paragraph{Baselines}

We compare the performance of LLMs to several baseline approaches: 
the best-performing feature-engineered approach (SVM)~\citep{Markov:2018}, a simple SVM approach with bag-of-words (BoW) features, BERT and GPT-2 approaches, with all scores directly cited from the original paper~\citep{lotfi-etal-2020-deep}. 

\paragraph{Closed-source LLMs} We rely on the results reported by \citet{zhang-2023} for GPT-3.5 (gpt-3.5-turbo) \citep{gpt3-brown:20} and GPT-4 (gpt-4-0613) \citep{openai-gpt4:23} on TOEFL11 and evaluate their performance on the ICLE-NLI dataset.

\paragraph{Open-source LLMs}
We conduct a comparative study of five recent open-source LLMs: LLaMA-2 (7B) \citep{llama2-2023-touvron}, LLaMA-3 (8B) \citep{llama3_2024}, Gemma (7B) \citep{gemma-2024-mesnard}, Mistral (7B) \citep{mistral-jiang-2023}, and Phi-3 (3.8B) \citep{phi3report}. While there is an ongoing debate surrounding the definition of `open-source' with the rise of LLMs \citep{liesenfeld_dingemanse_2024}, for the purpose of our experiments, we consider open-source models that are open in weights. 
Following~\citet{zhang-2023}, we carry out experiments in a zero-shot setup, both for the closed-set and open-set NLI tasks.

We run inference on the selected open-source LLMs using the same prompt as \citet{zhang-2023}, with the only difference that we instruct each model to respond using JSON dictionaries to restrict the model output to one L1 classification label. For the closed-set task, we include the set of possible L1s in the prompt. If the model classifies an L1 outside of the provided set of classes, we apply iterative prompting up to 5 times. 
For the open-set task, the prompt does not include a set of possible L1s. For both closed- and open-set tasks, we adapt the prompt to each model's prompt template. If a prediction cannot be extracted after 5 attempts, the predicted label is set to `other'. 
The prompts for closed-set and open-set tasks are provided in appendices \ref{appendix_closedsetprompts} and \ref{appendix_opensetprompts}, respectively. 
We use 4-bit quantized instruction-fine-tuned versions of the open-source LLMs when prompting out-of-the-box. 

In addition, we fine-tune the 4-bit quantized models on the TOEFL11 training set and under 5-fold cross-validation (5FCV) on ICLE-NLI\footnote{We used 5-fold cross-validation for a direct comparison with previous studies, e.g.,~\citep{lotfi-etal-2020-deep,Markov:2018}.} with QLoRA \citep{qlora-dettmers-2023}, 
using
the Hugging Face framework and Unsloth  library\footnote{https://unsloth.ai/}. 
The prompts used for fine-tuning are provided in Appendix \ref{appendix_finetuningprompts}.

\section{Results} \label{results}
Table~\ref{tab:results} shows the results in terms of classification accuracy (\%) for the baseline approaches and LLMs, both out-of-the-box and after fine-tuning, in closed-set and open-set settings. For open-source LLMs, we provide the average score and standard deviation over three runs to account for stochasticity in model inference and training. 

\subsection{Closed-Source LLMs}
We observe high accuracy scores on the ICLE-NLI dataset in our experiments using the GPT-3.5 and GPT-4 models. 
The results are in line with the state-of-the-art results on the TOEFL11 dataset reported in \citep{zhang-2023} and indicate that GPT-4 is able to identify the L1s of highly-proficient English learners both in closed-set and open-set classification experiments.  

\subsection{Open-Source LLMs Out-of-the-Box}
We note a surprisingly low performance of open-source LLMs when used out-of-the-box in a closed-set setting, with the exception of LLaMA-3 on ICLE-NLI. While GPT-4 achieves an accuracy of 91.7\% and 95.5\% on TOEFL11 and ICLE-NLI, respectively, the five open-source models obtain accuracy scores ranging between 13.6\% and 75.8\%. All open-source LLMs also perform worse than the baseline approaches, including the simple SVM model with BoW features. 
Some open-source LLMs tend to predict mostly one or two languages, e.g., Gemma predicting mostly French and LLaMA-2 mostly Chinese, which partially explains such low results. The large performance gap 
raises the concern that closed-source LLMs might have seen the NLI benchmark datasets in training. Additional research is required to explore the possibility of data leakage, e.g., by examining whether a model has memorized a given text using perplexity measurements \citep{carlini_extracting-2021}.

\subsection{Fine-Tuned Open-Source LLMs vs. Closed-Source LLMs}
The results indicate that the performance of open-source LLMs improves substantially after task-specific fine-tuning. 
Fine-tuned Gemma achieves an accuracy score of 90.3\% (±1.2) on the TOEFL11 dataset, nearly matching the results of GPT-4 as reported in \citep{zhang-2023}, and a near-perfect accuracy score of 96.6\% (±0.2) on the ICLE-NLI dataset, outperforming GPT-4 by 1.1\%. 
We also observe that the open-source models that perform best out-of-the-box do not necessarily demonstrate the best performance after fine-tuning. 

Previous studies comparing closed-source and fine-tuned open-source LLMs 
provide contradictory findings, with some researchers reporting a drop in accuracy of 16\% on sentiment classification for fine-tuned smaller language models (Flan-T5, 770M) compared to ChatGPT \citep{zhang-etal-2024-sentiment}, while others report that fine-tuned open-source LLMs (Qwen, 7B; LLaMA-3, 8B) outperform closed-source LLMs (GPT-3.5, GPT-4) on text classification tasks \citep{bucher-2024-finetunedsmallllmsstill, edwards-camacho-collados-2024-language, wang2024smartexpertsystemlarge}. The results presented in this study provide  evidence that fine-tuned open-source LLMs can achieve comparable performance to closed-source LLMs. 

We also observe that LLaMA-3 stands out with a high result on ICLE-NLI compared to TOEFL11. While out-of-the-box LLaMA-3 obtains 56.6\% accuracy on TOEFL11, it achieves a higher score of 75.8\% on ICLE-NLI. In addition, while all other open-source LLMs gain a large boost in performance after fine-tuning on both datasets, \mbox{LLaMA-3's} accuracy after fine-tuning on ICLE-NLI increases by 2.7 percentage points only. LLaMA-3's relatively high performance out-of-the-box and marginal performance boost after fine-tuning are inconsistent with the results for other open-source LLMs, possibly indicating that LLaMA-3 has seen the ICLE data in training. 

Comparing the confusion matrices for GPT-4 and fine-tuned Gemma, the best-performing closed-source and open-source LLMs (Appendix~\ref{appendix_confusion}), we note that both models tend to misclassify Hindi texts as Telugu in the TOEFL11 dataset. Hindi and Telugu have been considered a problematic language pair in previous studies on TOEFL11~\citep{malmasi-etal-2013-mq}. 
Fine-tuned Gemma has a tendency to misclassify Japanese essays as Korean. The high degree of confusion between Korean and Japanese has also been observed in previous research~\cite{markov-exploiting2022}. 
On ICLE-NLI, \mbox{GPT-4} erroneously classifies 
Bulgarian as Russian, both Slavic languages. Gemma misclassifies 14 Czech and Russian samples as Bulgarian. In line with previous research, we note that the 
confused L1s are either related through geographical location or belong to the same language family. 

\subsection{Closed-Set and Open-Set Settings}
We observe a drop in performance for most open-source LLMs from a closed-set to open-set setting, similarly to closed-source LLMs. 
Surprisingly, some of the models, i.e., GPT-3.5 and Phi-3, perform better in the open-set than in the closed-set setup. Further research is required to understand the reasons for this behaviour. 



\section{Conclusion} \label{conclusion}

We explored the performance of a variety of open-source LLMs for the NLI task. 
Our results indicate that open-source LLMs achieve lower performance than closed-source LLMs for this task when used out-of-the-box, while domain-specific fine-tuning of open-source LLMs allows these models to achieve comparable results to the proprietary LLMs, such as GPT-4, on the benchmark TOEFL11 and ICLE-NLI datasets. We believe that our work opens up avenues
for future research on LLM-based Native Language Identification. Future research could explore few-shot prompting and  different prompt variations as a way to potentially boost the performance of open-source LLMs. 

\section*{Limitations} \label{limit}

\paragraph{Multilingual NLI} Our study focuses purely on native language identification in English, which is the most well-studied L2 in the NLI task \citep{goswami-etal-2024-native}. It would be interesting to explore whether the high performance of LLMs on NLI holds for L2s other than English.

\paragraph{Fine-tuned LLMs in cross-corpus setting} While fine-tuning 
 drastically improves the performance of open-source LLMs, the prerequisite of fine-tuning for optimal performance is a disadvantage for open-source LLMs compared to closed-source LLMs. Previous research has shown that NLI models suffer from performance degradation in a cross-corpus setting, and thus cannot be applied directly to different corpora \citep{markov-exploiting2022, malmasi-dras-2015-large}. Future research could explore the use of fine-tuned open-source LLMs for NLI in a cross-corpus setup.  

\paragraph{Defining open-source LLMs} More broadly, in our study, we define open-source and closed-source relatively loosely, treating the terms `open' and `closed' as a binary feature to perform a comparative analysis between open-source and closed-source LLMs for NLI. However, there are various dimensions of openness, as a model release involves different components ranging from the 
disclosure of training datasets to model access \citep{solaiman2023gradientgenerativeairelease, liesenfeld_dingemanse_2024}. Most providers of proclaimed open-source LLMs release little to no information regarding their training data and procedure, despite framing them as being open-source. 
In turn, it is difficult to determine whether an open-source model's performance can be attributed to the model's learning or possible data contamination. The lack of insights into the training data of proclaimed open-source LLMs also hindered our evaluation of LlaMA-3 on the ICLE-NLI dataset. 


\bibliography{coling_latex}

\appendix

\section{Hyperparameters and Computation Time}
We fine-tuned the open-source LLMs with the following hyperparameters: a learning rate of 1e-4, batch size of 16, 3 epochs, and optimization via AdamW optimizer. The experiments were conducted on Google Colaboratory Pro with the A100 GPU (40 GB RAM). The models were loaded with 4-bit NF-quantization and QLoRA adapters were added and fine-tuned using the bitsandbytes library\footnote{https://huggingface.co/docs/bitsandbytes}. The total computation time was roughly 120 hours. Total emissions are estimated to be 17.1 kgCO$_2$eq of which 100\% was directly offset by the cloud provider\footnote{Estimations were conducted using the \href{https://mlco2.github.io/impact/}{Machine Learning Impact calculator} \citep{lacoste2019quantifying}.}.

\section{Confusion Matrices}

The confusion matrices are provided in Figure~\ref{fig:TOEFL-ICLE}.
\label{appendix_confusion}
\begin{figure*}[!htb]
\centering
\includegraphics[width=0.42\textwidth]{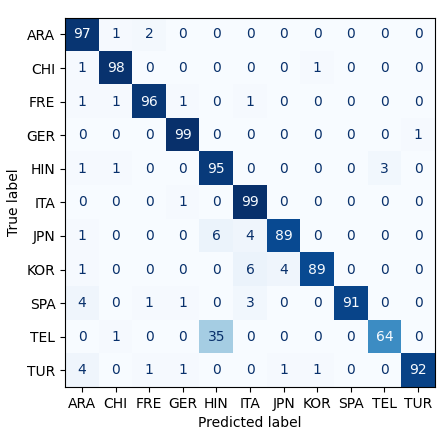}
\includegraphics[width=0.42\textwidth]{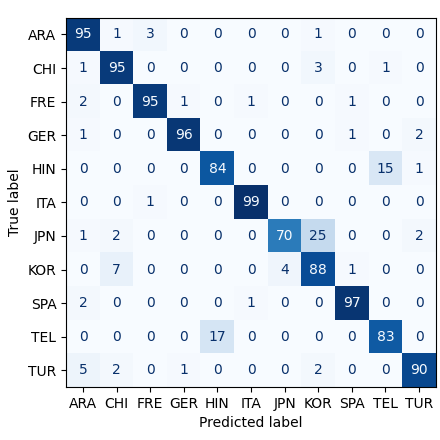}  \\
\includegraphics[width=0.42\textwidth]{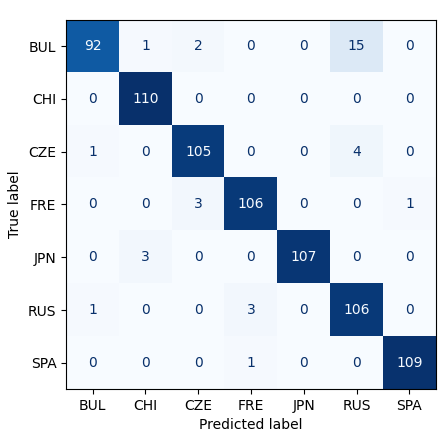} 
\includegraphics[width=0.42\textwidth]{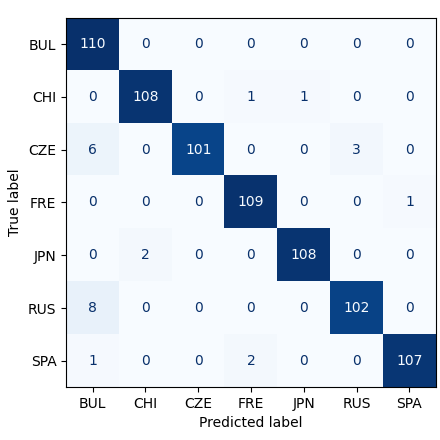}
\caption{Confusion matrices for GPT-4 on TOEFL~\citep{zhang-2023} (top left), Gemma (7B) (fine-tuned) on TOEFL (top right).
GPT-4 on ICLE-NLI (bottom left), Gemma (7B) (fine-tuned) on ICLE-NLI (bottom right).}
\label{fig:TOEFL-ICLE}
\end{figure*}

\section{LLM Prompts}
\label{sec:appendix}

\subsection{Closed-Set Prompts}
\label{appendix_closedsetprompts}
For the closed-set experiments on the TOEFL11 dataset, we used the prompts below. For ICLE-NLI, we used exactly the same prompts, with the only difference being the set of possible L1s covered in the dataset.

\begin{tcolorbox}
\small
  You are a forensic linguistics expert that reads English texts written by non-native authors to classify the native language of the author as one of:\\\\
  ``ARA'': Arabic\\
  ``CHI'': Chinese\\
  ``FRE'': French\\
  ``GER'': German\\
  ``HIN'': Hindi\\
  ``ITA'': Italian\\
  ``JPN'': Japanese\\
  ``KOR'': Korean\\
  ``SPA'': Spanish\\
  ``TEL'': Telugu\\
  ``TUR'': Turkish\\
  Use clues such as spelling errors, word choice, syntactic patterns, and grammatical errors to decide on the native language of the author.\\
  
DO NOT USE ANY OTHER CLASS.\\
IMPORTANT: Do not classify any input as ``ENG'' (English). English is an invalid choice.\\

Valid output formats:\\
Class: ``ARA'', \\
Class: ``CHI'', \\
Class: ``FRE'', \\
Class: ``GER'' \\

You ONLY respond in JSON files. The expected output from you is: json \{``native\_lang'': The chosen class, ARA, CHI, FRE, GER, HIN, ITA, JPN, KOR, SPA, TEL, or TUR\}

\end{tcolorbox}
When possible, the prompt above was entered as a System prompt. If the system role was not supported by the prompt formatter, the prompt was entered as part of the User prompt. We input the given text and used the prompt below as a User prompt: 
\begin{tcolorbox}
\small
$<$TOEFL11 ESSAY TEXT$>$ \\
Classify the text above as one of ARA, CHI, FRE, GER, HIN, ITA, JPN, KOR, SPA, TEL, or TUR. Do not output any other class - do NOT choose ``ENG'' (English). What is the closest native language of the author of this English text from the given list?
\end{tcolorbox}

In the closed-set experiments, if the L1 was incorrectly predicted as English, we prompted the model again using the prompt below:
\begin{tcolorbox}
\small
    You previously mistakenly predicted this text as ``ENG'' (English). The class is NOT English. Please classify the native language of the author of the text again.
\end{tcolorbox}

If we were unable to parse the prediction or the predicted L1 was not in the set of possible classes, we prompted the model again. For the TOEFL11 experiments, we used the prompt below:
\begin{tcolorbox}
\small
    Your classification is not in the list of possible languages.\\
Please try again and choose only one of the following classes:
ARA, CHI, FRE, GER, HIN, ITA, JPN, KOR, SPA, TEL, or TUR
\end{tcolorbox}

\subsection{Open-Set Prompts}
\label{appendix_opensetprompts}
For the open-set experiments, we used the prompt below as an input prompt for all the models: 
\begin{tcolorbox}
\small
    You are a forensic linguistics expert that reads texts written by non-native authors in order to identify their native language. \\Analyze each text and identify the native language of the author.\\Use clues such as spelling errors, word choice, syntactic patterns, and grammatical errors to decide.\\\\
    You ONLY respond in JSON files. The expected output from you has to be: ``json \{``native\_lang'': ``''\}''
\end{tcolorbox}

If the predicted L1 could not be extracted from the generated output, we used the prompt below to apply iterative prompting to get a valid prediction: 
\begin{tcolorbox}
\small
    Your previous classification was not in the correct format. Please only respond in the following JSON format:\\``json \{``native\_lang'': ``''\}''
\end{tcolorbox}

\subsection{Fine-Tuning Prompts}
\label{appendix_finetuningprompts}
We used the following prompt for the fine-tuning experiments: 

\begin{tcolorbox}
\small
\#\#\# Instruction: \\
You are a forensic linguistics expert that reads English texts written by non-native authors to classify the native language of the author as one of:\\\\
  ``ARA'': Arabic\\
  ``CHI'': Chinese\\
  ``FRE'': French\\
  ``GER'': German\\
  ``HIN'': Hindi\\
  ``ITA'': Italian\\
  ``JPN'': Japanese\\
  ``KOR'': Korean\\
  ``SPA'': Spanish\\
  ``TEL'': Telugu\\
  ``TUR'': Turkish\\
Use clues such as spelling errors, word choice, syntactic patterns, and grammatical errors to decide on the native language of the author.\\

DO NOT USE ANY OTHER CLASS.\\
IMPORTANT: Do not classify any input as ``ENG'' (English). English is an invalid choice.\\

Valid output formats:\\
Class: ``ARA'', \\
Class: ``CHI'', \\
Class: ``FRE'', \\
Class: ``GER''\\ 

Classify the text below as one of ARA, CHI, FRE, GER, HIN, ITA, JPN, KOR, SPA, TEL, or TUR. Do not output any other class - do NOT choose ``ENG'' (English). What is the closest native language of the author of this English text from the given list?\\

\#\#\# Input:\\
$<$TOEFL11 ESSAY TEXT$>$\\

\#\#\# Response:\\
$<$L1 LABEL$>$
\end{tcolorbox}

\end{document}